Abstract: This paper presents the design of an artificial vision system prototype for automatic inspection and monitoring of objects over a conveyor belt and using a Smart camera 2D BOA-INS. The prototype consists of a conveyor belt and an embedded system based on an Arduino Mega card for system control, and it has as main peripherals the smart camera, a direct current motor, a photoelectric sensor, LED illumination and LEDs indicating the status (good or defect) of each evaluated object. The application of the prototype is for educational purposes, so that undergraduate, master and diploma students can simulate a continuous production line, controlled by an embedded system, and perform quality control by monitoring through a visual system and a personal computer. This allows implementing the topics of embedded systems, artificial vision, artificial intelligence, pattern recognition, automatic control, as well as automation of real processes.

# Vision System Prototype for Inspection and Monitoring with a Smart Camera

E. Hernández-Molina, B. Ojeda-Magaña, J. G. Robledo-Hernández and R. Ruelas

*Abstract—* **This paper presents the design of an artificial vision system prototype for automatic inspection and monitoring of objects over a conveyor belt and using a Smart camera 2D BOA-INS. The prototype consists of a conveyor belt and an embedded system based on an Arduino Mega card for system control, and it has as main peripherals the smart camera, a direct current motor, a photoelectric sensor, LED illumination and LEDs indicating the status (good or defect) of each evaluated object. The application of the prototype is for educational purposes, so that undergraduate, master and diploma students can simulate a continuous production line, controlled by an embedded system, and perform quality control by monitoring through a visual system and a personal computer. This allows implementing the topics of embedded systems, artificial vision, artificial intelligence, pattern recognition, automatic control, as well as automation of real processes.**

## I. INTRODUCCIÓN

LA INSPECCIÓN de productos en la industria mediante Sistemas de Visión Artificial (SVA) ha avanzado enormemente durante los últimos años. Esto ha conducido a la necesidad de desarrollar la toma de decisiones automática que permita realizar tareas tales como detección de condiciones del entorno por cuestiones de seguridad, clasificación de productos, detección de artículos defectuosos, medición de piezas, etc. a partir de los resultados del procesamiento de imágenes digitales [1] o del estado de diferentes tipos de sensores, pudiendo calcular así datos estadísticos de la producción, entre otros [2]. Esto ha sido posible gracias al desarrollo de la microelectrónica y al desarrollo de Sistemas Embebidos (SE), que permiten crear desde sistemas muy simples hasta soluciones totalmente integradas, aún si están espacialmente distantes. Por ejemplo, es así como los SVA han ido sustituyendo poco a poco la inspección visual humana, debido a que no se ven afectados por factores psicológicos, tales como la fatiga o los hábitos [3], es más, los SVA ofrecen mayor velocidad de operación, precisión y repetibilidad. En una línea de producción, los SVA pueden inspeccionar cientos o miles de piezas por minuto de manera confiable y repetida, superando ampliamente las capacidades de la inspección humana.

Por otra parte, los SE y los SVA también son cada vez más comunes en teléfonos móviles, domótica, seguridad, robótica, entre otros. Este crecimiento ha favorecido el desarrollo tecnológico, abaratado además los costos y mejorado las prestaciones de adquisición y procesamiento de datos, información, imágenes y video, así como la funcionalidad del software. De igual manera, los sistemas embebidos están desempeñando un papel cada vez más importante en el internet de las cosas, aprendizaje profundo en grandes volúmenes de datos, así como en la producción de información y procesamiento en la nube. Las cámaras inteligentes como la BOA-INS (BOA-INSpect) [4], por ejemplo, tienen un sistema embebido específico para visión artificial y que, junto con una aplicación de software, permiten disponer de una interfaz gráfica de usuario (GUI en inglés), en donde se presentan visualmente los componentes y los algoritmos del sistema de visión artificial; la programación gráfica proporciona facilidad y flexibilidad en la solución de las tareas de visión [2].

En este trabajo se presenta el diseño de un prototipo con el fin crear una interacción de alumnos y usuarios con entornos reales y multidisciplinarios, que permitan profundizar en su aprendizaje, tanto en sistemas embebidos, como en SVA. Por lo tanto, se desarrolló un SE con el propósito de generar una solución para aplicaciones de visión artificial, para lo cual se utilizaron, básicamente, dos sistemas: un sistema embebido basado en una tarjeta Arduino Mega2560 para el control total del prototipo y una cámara inteligente BOA-INS, también controlada por el SE. El primero dedicado a la interacción con el usuario, al control de periféricos, a la activación de indicadores, al registro de eventos y a la manipulación de otros sistemas y sensores, mientras que el segundo está dedicado a la adquisición y procesamiento de las imágenes digitales.

En la actualidad es cada vez más común el desarrollo de prototipos con el fin de comprender, aprender o entrenar en determinados campos del conocimiento o técnicas específicas. Por ejemplo, en [5] se presenta "un robot que juega a las damas" el cual integra tecnologías como; la programación de robots, la visión artificial y la inteligencia artificial. Dicho robot ha sido realizado con fines educativos y ha permitido elevar la motivación de los estudiantes, además de desarrollar competencias transversales como el trabajo en equipo, la planificación y la creatividad. El robot es utilizado como material didáctico en clases de robótica.

Por otra parte, en [6] desarrollaron un laboratorio virtual para el procesamiento de imágenes digitales, el cual permite

This paper was submitted for review on September 13th, 2019. This work was supported in part by Universidad de Guadalajara under Grant Fortalecimiento del UDG-CA-379 2019, and in part by SEP with the incorporation of a new PTC through Grant UDG-NPTC-1461.

The authors are with the Departamento de Ingeniería de Proyectos, CUCEI, Universidad de Guadalajara, José Guadalupe Zuno # 48, Los Belenes, CO 45150 Zapopan, MEXICO (e-mail: ehm_16@hotmail.com, benojed@hotmail.com, jguadalupe.robledo@academicos.udg.mx, ruben.ruelas@academicos.udg.mx).

degradar las imágenes en rango y azimuth (ejes *x, y*), aplicar ruido, e implementar diversos algoritmos de reconstrucción y post-procesamiento de imágenes, permitiendo a los estudiantes analizar y comprender mejor el conjunto de algoritmos utilizados en la asignatauras de Visión por computadora y Procesamiento digital de señales, y ha tenido una buena aceptación por parte de los estudiantes. En [7] se presenta otro caso que consiste en un clasificador de imágenes de frutas (manzanas rojas, manzanas verdes, plátanos y naranjas), para ello se extraén regiones de interés, o regiones donde se encuentran las frutas, de cada una de las imágenes y se extraén características de la imágenes como: color, forma y textura; posteriormente, y basado en un algoritmo de inteligencia artificial, clasifican automáticamente las frutas.

Este tipo de desarrollos no se ha limitado al entorno educativo y también podemos encontrar un gran número de investigaciones enfocadas a aplicaciones industriales como se puede ver en [8], donde se trabaja en procesos de inspección de calidad y se determina, a través de un sistema de visión artificial, sí es correcto el número de frascos de mayonesa dentro de una caja de cartón, o sí a los envases de mayonesa les hace falta la tapa; este proceso se realiza en tiempo real. En [9] se hace uso de un sistema de visión artificial y de redes neuronales artificiales, con lo cual realizan un aprendizaje compensado con ruido, para realizar la clasificación automática de fresas en la industria alimentaria. Este sistema es capaz de identificar: las fresas de máxima calidad; las fresas de consumo, que son visual y subjetivamente agradables para los humanos; las fresas que pueden ser usadas como materia prima industrial, y que son poco agradables para el consumo humano; y los cuerpos extraños (hojas, palitos, etc.).

Un sistema embebido es un dispositivo programable dedicado, generalmente, a una sola función y que puede actuar solo o como parte de un sistema más grande para obtener sistemas que operan en tiempo real, de forma reactiva o en interacción con el entorno. Así, estos son específicos a cada aplicación y pueden verse limitados por su máxima frecuencia de operación, la cantidad de memoria disponible y la disipación de potencia. Sin embargo, estos deben ser muy confiables cuando se desarrollan para aplicaciones críticas. Como se menciona en [10], los sistemas embebidos permiten una enseñanza más orientada a los problemas y organizada en forma de proyectos en las disciplinas involucradas favoreciendo un desarrollo con conocimientos más profundos y competencias probadas.

El prototipo que se presenta en este trabajo permite que los estudiantes exploren el desarrollo de sistemas embebidos desde el punto de vista de la función, que en este caso corresponde a una aplicación de visión artificial basada en una cámara inteligente, siendo el SE el que controla, coordina y registra la información y cálculos estadísticos. Además, también permite el desarrollo de OBD (On-Board Diagnostic) tanto para la parte operativa como para funciones de seguridad del usuario y del prototipo. No obstante, el SE está basado en un microcontrolador de 8 bits y por ello puede ser considerado de poca complejidad en una clasificación de 3 categorías (poca, mediana y alta o sofisticada complejidad).

A diferencia de las aplicaciones típicas de los SE, que por lo general tratan de aplicaciones de control y temas relacionados, en esta aplicación el SE se ha desarrollado para una aplicación de visión artificial. Así, el estudiante, como desarrollador o como usuario, tiene la oportunidad de aprender herramientas de hardware y software de visión artificial para la inspección de piezas y todas las implicaciones que esto conlleva para tener el máximo desempeño del sistema en el reconocimiento correcto de objetos entre los que podemos encontrar, por ejemplo, la velocidad de la banda transportadora, la sintonía del controlador PID y la iluminación led, entre otros, de tal forma que el usuario experimente un entorno cercano a uno de tipo industrial.

El resto del trabajo está organizado de la siguiente manera: los materiales del prototipo y el sistema embebido se describen en la Sección 2; la implementación de 3 casos de estudio además del análisis de los resultados e impacto obtenidos se describen en la Sección 3; y, finalmente, las principales conclusiones de este estudio se presentan en la Sección 4.

## II. DESCRIPCION DE LA ESTACION

En esta sección se describen a detalle los componentes que conforman el prototipo; como se puede ver en la Figura 1 y en la Figura 2. Este prototipo ha sido desarrollado principalmente para la asignatura de visión por computadora de la Maestría en Proyectos Tecnológicos y para el Diplomado en Automatización y Control Industrial, así como proyectos de licenciatura de la Universidad de Guadalajara. Entre los contenidos de los programas se describen los procedimientos para el desarrollo de sistemas embebidos y el desarrollo de sistemas inteligentes, las metodologías de procesamiento de imágenes tales como: detección de borde, algoritmos de segmentación, método de Otsu, morfología matemática, filtrado espacial, entre otros [11], así como el uso de técnicas de inteligencia artificial como, la lógica difusa [12], los algoritmos genéticos [13], redes de aprendizaje profundo, entre otros, para resolver problemas de visión de alto nivel que se presentan en la industria.

### A. Componentes del prototipo

El prototipo cuenta con los siguientes componentes para su funcionamiento (ver Figura 1 y Figura 2):

1. Elementos mecánicos.
2. Banda transportadora.
3. Sistema embebido.
4. Cámara inteligente.
5. Lámparas de iluminación LED.
6. Sensor fotoeléctrico.
7. Motor de corriente directa.
8. Monitoreo por PC.
9. Fuente de alimentación de 12 $V_{cd}$.

*Elementos mecánicos*

El diseño mecánico del prototipo (Figura 1) se realizó con el CAD SolidWorks-2018, el cual permite tomar decisiones de la fabricación mecánica y del ensamble. Adicionalmente, el software también permite la simulación del prototipo en 3D previo a su fabricación.

Para la fabricación de las piezas se utilizó aluminio y se trabajó en máquinas herramienta (torno y fresadora). Para el mecanismo de la banda transportadora se utilizaron rodamientos de bolas y un riel de rodillos plásticos.

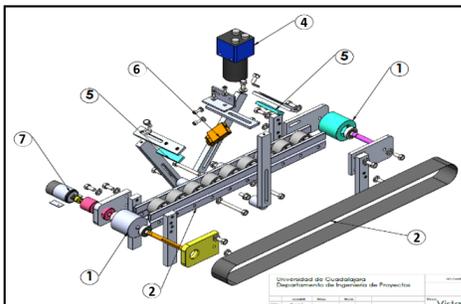

Figura 1. Dibujo CAD de las piezas que conforman la parte mecánica del prototipo.

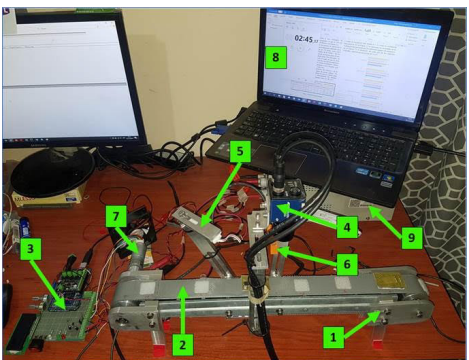

Figura 2. Componentes del prototipo de inspección por visión artificial.

*Banda Transportadora*

La banda transportadora del prototipo, es un diseño de banda horizontal que tiene una longitud total de 50 cm y 4 cm de ancho, fabricada de PVC. La cual es tensionada por sus extremos con el fin de recibir el movimiento radial de la polea motriz, acoplada a un motor de corriente directa. El diseño de la banda simula una producción industrial real. En cuanto a su funcionamiento, la banda trabaja continuamente por lo que el sistema embebido cuenta con un sensor para detectar la presencia de un objeto y que la cámara le tome la imagen al vuelo. La velocidad máxima de la banda es de 67 cm/s y el número máximo de objetos que se detectan es de 390 piezas por minuto.

*Sistema embebido*

El sistema embebido del prototipo consta de una tarjeta Arduino Mega2560 de 8 bits, frecuencia de reloj de 16 MHz, memoria flash de 256 kB de los cuales 8 kB se usan para el arranque y voltaje de operación de 5 V. Además, el SE cuenta con un LCD, reguladores de voltaje, leds, potenciómetros y una tarjeta controladora de motores de corriente directa (*shield*) con capacidad de carga de 2 A. Asimismo, para obtener condiciones de operación más regulares de la banda transportadora, se implementó un algoritmo de control PID (Proporcional-Integral-Derivativo). Este tipo de control fue necesario para garantizar una velocidad estable y que los objetos o piezas siempre estuvieran dentro del área de inspección. La aplicación de visión artificial requiere al menos de la especificación de los siguientes parámetros:

- *Set Point* para la velocidad de la banda transportadora.
- *Set Point* para la intensidad luminosa.
- Tiempo de espera para la estabilización de la velocidad de la banda transportadora.
- Lectura del codificador del motor.
- Monitoreo del sensor fotoeléctrico.
- Pulso de disparo (*trigger*) hacia la cámara BOA.
- Pantalla LCD para monitoreo de la velocidad del motor.

El SE cuenta con dos indicadores relacionados al resultado de la evaluación de cada imagen, un color verde para las piezas identificadas como correctas y un led rojo para las incorrectas. Aunque aquí sólo se usa como indicador, este último se puede usar para activar un actuador y eliminar objetos de la banda transportadora por no cumplir con la calidad esperada.

*Cámara Inteligente BOA*

La adquisición de la imagen digital es la parte más importante en un SVA, debido a que representa la materia prima para el procesamiento de la imagen adquirida, y lograr obtener un reconocimiento de la falla en la imagen del objeto [14]. Para lograr esto debe existir un trabajo coordinado del control electrónico, los elementos mecánicos, la iluminación y la cámara inteligente. Aunque el control y los datos estadísticos los lleva a cabo el sistema embebido, el prototipo también usa una PC para monitoreo, visualización y registro de datos generados por la cámara digital.

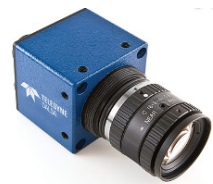

Figura 3. Cámara inteligente BOA.

La cámara inteligente utilizada para este trabajo es de la marca Teledyne Dalsa, modelo BVS-0640C-INS-UNI (ver Figura 3), la cual tiene una resolución de 640X480 pixeles a color con un tamaño de pixel de 7,4 μm, una velocidad máxima de operación de 60 fps y una interfaz para conexión a ethernet.

La cámara inteligente BOA, es un sistema de visión altamente integrado en un formato compacto, diseñado específicamente para su uso en aplicaciones industriales. Cuenta con un software embebido, que para el modelo de cámara es el software iNspect, el cual combina la facilidad de uso, con un conjunto de herramientas y capacidades que se pueden utilizar simultáneamente en las aplicaciones de inspección en procesos industriales [15]. El software iNspect ofrece 25 herramientas de interfaz gráfica para la inspección de piezas u objetos, según sea el problema industrial (Figura 4) [4]. Esta cámara inteligente tiene comunicación con la computadora personal mediante una interfaz Ethernet 100 Base-T de bajo costo. Además, la cámara BOA tiene integradas señales Entrada/Salida que le permiten activar dispositivos externos como un Controlador Lógico Programable (PLC por sus siglas en inglés) y otros actuadores [16].

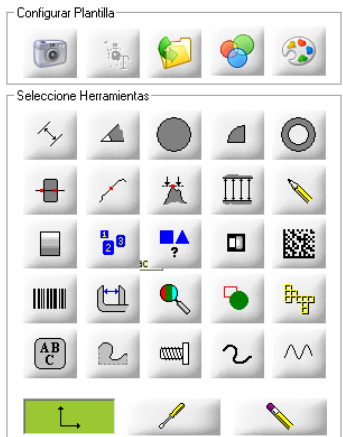

Figura 4. Menú de herramientas de iNspect Express.

*Lámparas de iluminación LED*

La iluminación es una parte fundamental de la visión artificial [17]. Los objetos o piezas a identificar deben ser iluminados adecuadamente ya que la inspección se realiza a través de una imagen digital, en donde las sombras o los colores pueden provocar falsos negativos o falsos positivos, es decir, dejar de identificar parte de los objetos o piezas o, por el contrario, identificar partes de más. Al prototipo se le instalaron 2 módulos de iluminación LED a los lados de la cámara. Cada módulo cuenta con tres LEDs tipo 5730 SMD y tiene como características: color de la iluminación blanco frio (8,000 k), flujo luminoso de 120 lúmenes, ángulo de apertura de 120 °, voltaje de operación de 12 $V_{cd}$ y consumo de potencia de 12 W. Estas lámparas están instaladas con mecanismos independientes que proporcionan 4 ejes de movimiento para facilitar el ajuste de la iluminación de acuerdo al tipo de objeto, velocidad de trabajo de la banda, efectos del entorno, etc. Adicionalmente, la intensidad luminosa puede ser ajustada desde el sistema embebido. De esta manera, el diseño es amigable por la facilidad de ajuste con respecto a los objetos o piezas a inspeccionar y las condiciones en que se lleva a cabo la prueba.

*Sensor fotoeléctrico*

El sensor fotoeléctrico utilizado en este proyecto es un efector 200 de IFM, correspondiente a un sensor de reflexión difusa para la detección sin contacto de objetos y materiales, el cual indica la presencia de éstos mediante una señal conmutada. La función de este sensor es detectar cuándo un objeto o pieza está en el área de inspección para enviar una señal de confirmación a la tarjeta Arduino. A su vez, la tarjeta de control envía un pulso de gatillo (*Trigger*) hacia la cámara, que mantiene durante varios microsegundos. Con esta señal la cámara hace una toma del objeto transportado por la banda e inicia el procesamiento de la imagen. Posteriormente se determina si la pieza o el objeto tiene defecto o falla, o si está normal.

*Motor de corriente directa*

El motor utilizado es un motor de corriente directa con un voltaje de operación de 6 a 12 $V_{cd}$, con reductor de engranes de 1:34 y un codificador (encoder) tipo cuadratura, 5,2 kg*cm/110 rpm/3.1 W/1.10 A. El codificador está acoplado después de la caja de engranes y genera 748 pulsos por revolución. El consumo de energía del motor es elevado y por ello se ha utilizado una fuente de voltaje externa [18] con un voltaje de salida de 12 $V_{cd}$.

*B. Metodología de funcionamiento.*

En la Figura 5 se muestra el diagrama de flujo de operación del prototipo, el cual está divido en dos partes: a) en color naranja lo correspondiente al control de periféricos y b) en color verde lo que se refiere a las acciones que realiza el sistema para la caracterización de objetos.

En primera instancia, el sistema embebido ejecuta el controlador PID para el control de la velocidad. Posteriormente, el algoritmo utiliza el tiempo de estabilización del control PID antes de realizar la verificación de los objetos o piezas. De esta manera, una vez que se encuentra en estado estable, el sistema de control habilita el sensor fotoeléctrico, cuya función es la de proporcionar la señal de disparo y captura de la imagen cada vez que se detecta un objeto.

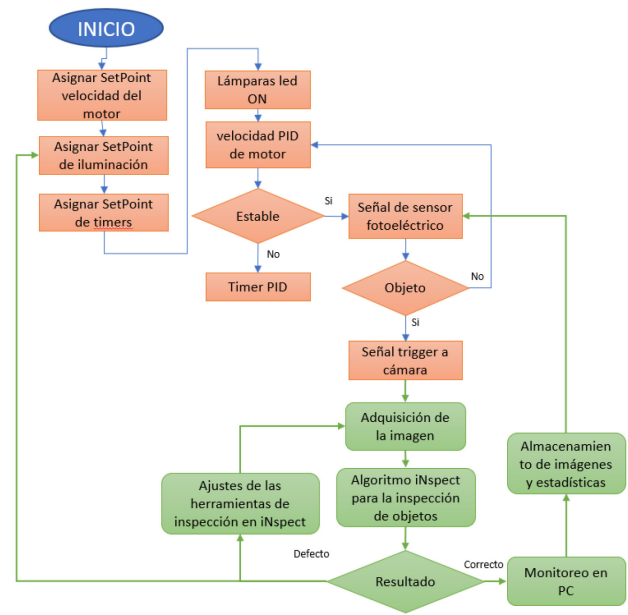

Figura 5. Diagrama de flujo sobre el funcionamiento general del prototipo.

Por otro lado, una vez que la cámara inteligente captura una imagen de un objeto transportado por la banda, inmediatamente se ejecutan uno o varios algoritmos de inspección del objeto que posteriormente toman la decisión respecto a si el objeto en la imagen cumple o no los criterios de calidad. Enseguida se emite el mensaje correspondiente en el monitor de la PC. Simultáneamente, el algoritmo de inspección genera una colección de datos estadísticos y los hace visibles en el monitor.

Para los ajustes de trabajo iniciales, se requiere cumplir con las siguientes actividades en el orden descrito en la Figura 6 y en la Figura 7.

```
 8 //Variables globales
 9 double Setpoint=200
10 byte NivelLuz=255;
11 short Tespera=6000;
12 short Timetrig=2000;
```

Figura 6. Definición de parámetros del sistema de control.

1. Establecer la comunicación entre cámara inteligente, sistema de control y PC.

2. Fijar objetos que no tienen defectos en la banda transportadora.

3. Ajustar los parámetros del sistema de control (Arduino Mega).

4. Seleccionar el elemento de referencia en el software iNspect para verificar cada uno de los componentes.

5. Iniciar la operación del sistema.

Figura 7. Listado de actividades iniciales.

### III. CASOS DE ESTUDIO Y RESULTADOS

Para determinar la funcionabilidad del prototipo se hizo la selección de 3 casos de estudio muy comunes en la industria. Por lo tanto, se simularon estos procesos utilizando diferentes productos entre los cuales se presentan productos con defectos o fallas. De igual manera, además de hacer un análisis de los resultados, se llevó a cabo un estudio sobre la aceptación o no del prototipo por parte de diferentes grupos de usuarios.

*A. Casos de estudio*

Los casos de estudio son:
a. Inspección de la correcta impresión del código de barras en envases de polipropileno de alta densidad impresos por serigrafía.
b. Inspección de la presencia de resistencias eléctricas en circuitos impresos o PCBs (*Printed Circuit Boards*).
c. Inspección de la presencia y detección del tipo de tornillos entre dos casos posibles.

El primer caso de estudio, impresión de envases de plástico por serigrafía, tiene una gran demanda en la industria. La producción diaria de este tipo de productos, que se produce por lotes, se cuentan por cientos o por miles por día, por lo que es relativamente común la presencia de defectos o fallas en la impresión; ver Figura 8 en donde se observan ejemplos de códigos de barras mal impresos.

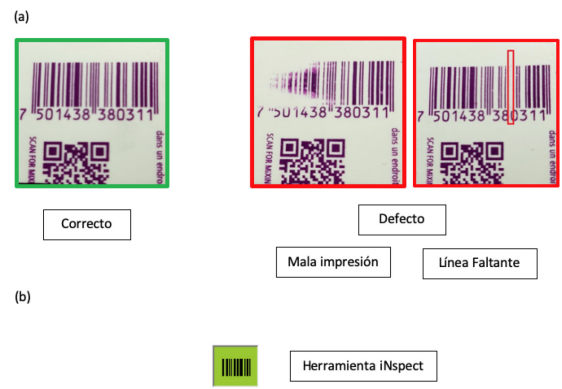

Figura 8. Ejemplos de fallas para el primer caso de estudio, a) Códigos de barra impresos y b) Herramienta utilizada.

Para el segundo caso de estudio se seleccionó la inspección de PCBs para determinar automáticamente la presencia o ausencia de un elemento, una resistencia eléctrica en este caso, en circuitos impresos. Ver Figura 9.

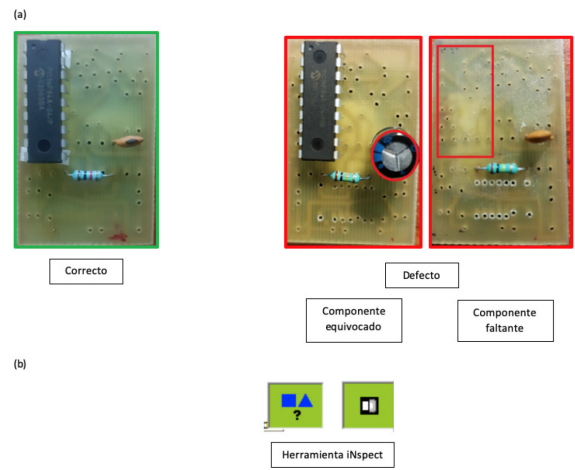

Figura 9. Ejemplos de fallas para el segundo caso de estudio, a) Tarjetas electrónicas y b) Herramientas utilizadas.

En el tercer caso de estudio se seleccionó un proceso de verificación de la presencia y tipo de tornillos presentes. En la Figura 10 se observan tornillos conformes a la especificación, es decir, 2 tornillos con cabeza hexagonal, mientras que la otra imagen muestra un tornillo con cabeza redonda que debe ser descartado al ser identificado como defecto con respecto a los elementos de referencia de iNspect.

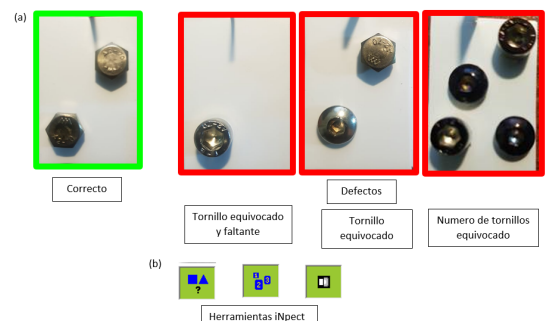

Figura 10. Ejemplo de falla para el tercer caso de estudio, a) Tornillos y b) Herramientas utilizadas.

## B. Resultados

Para la experimentación se seleccionaron 10 objetos para cada caso de estudio, denominados aquí como A, B y C, de los cuales se utilizaron 5 objetos considerados como correctos, o sin defectos, y 5 incorrectos o con defectos. El desempeño de la clasificación de los objetos se puede verificar con los indicadores de proporción de verdaderos positivos (VP), con el cual se puede medir la efectividad del clasificador (ver ecuación (1)), y con la proporción de verdaderos negativos (VN) que nos muestra la especificidad del clasificador (ver ecuación (2)). Frente a este escenario, la conclusión, si todo funciona correctamente, es que el 50% de los objetos PASA y el otro 50% NO PASA (ver Figura 11).

$$Sensibilidad = \frac{VP}{(VP + FN)} * 100\ \% \quad (1)$$

$$Especificidad = \frac{VN}{(VN + FP)} * 100\ \% \quad (2)$$

Después de que el prototipo se ha puesto en operación a velocidad máxima y conforme al procedimiento señalado en la Figura 6, éste se ha dejado trabajar por 50 ciclos; todos estos datos y los resultados están reportados en la Tabla 1. Los porcentajes de desempeño nos indican que tenemos un área de oportunidad para ajustar los parámetros y reducir los errores al máximo. Ahora, estos resultados son normales conforme a los desafíos que presentan los objetos utilizados en cada caso.

La versatilidad de la parte mecánica, del control y de la cámara es de gran ayuda para lograr un mejor trabajo. No obstante, también es un reto encontrar los ajustes ideales para cada caso, ya que no aplican exactamente las mismas condiciones de solución en los diferentes casos (A, B y C).

| Caso | Objetos inspeccionados | Tiempo de prueba | Resultados | | Desempeño | |
|---|---|---|---|---|---|---|
| | | | Falsos positivos | Falsos negativos | Sensibilidad | Especificidad |
| A | 3268 | 23'42" | 0 | 1 | 99.96% | 100.00% |
| B | 2925 | 31'10" | 2 | 53 | 96.09% | 99.87% |
| C | 4682 | 61'02" | 0 | 255 | 89.10% | 100.00% |

Tabla 1. Resultados experimentales.

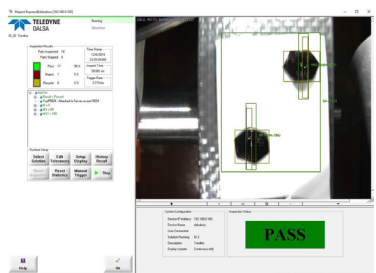

Figura 11. Pantalla de prueba funcional.

## C. Impacto obtenido

Con el fin de evaluar la percepción, utilidad y grado de aceptación del prototipo por diferentes grupos de usuarios (visitantes al departamento, diplomados, prestadores de servicio, estudiantes de maestría y maestros), se realizó un estudio estadístico con grupos de muestras de tamaño similar para los periodos académicos 2018 B y 2019 A. Para llevar a cabo esta evaluación se tomaron en cuenta 4 aspectos: 1) Facilidad de uso, 2) Claridad en configuración y ajustes, 3) Facilidad de programación y 4) Aplicaciones futuras del prototipo. El universo encuestado estuvo formado por 100 personas divididos en 4 grupos de interés a) visitantes al laboratorio y estudiantes del diplomado, b) estudiantes de maestría, c) profesores de la institución y d) prestadores de servicio social de las carreras de ingeniería mecánica eléctrica y de ingeniería en comunicaciones y electrónica.

Para la evaluación se les pidió a los participantes utilizar una escala del 0 al 10 para cada uno de los aspectos, considerando 0 como la evaluación más negativa y 10 como la evaluación más positiva. La Figura 12 muestra la segmentación del universo de 100 encuestados.

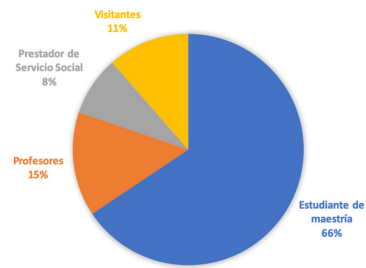

Figura 12. Cantidad de personas encuestadas por categorías.

En la Figura 13, se muestran los resultados de las opiniones respecto al cuestionamiento relacionado con la facilidad de uso del prototipo. Se observa que la mayoría de las opiniones se concentran entre estar totalmente de acuerdo (17), muy de acuerdo (33) y de acuerdo (29) en relación a la facilidad de uso del prototipo. Por otro lado, se observan 20 resultados en las franjas de valores del 5 al 7 lo cual indica que los encuestados tienen opiniones neutras al respecto. También se observa una opinión que está en desacuerdo. El valor medio para este aspecto es de 9.24 y la desviación estándar de 1.46.

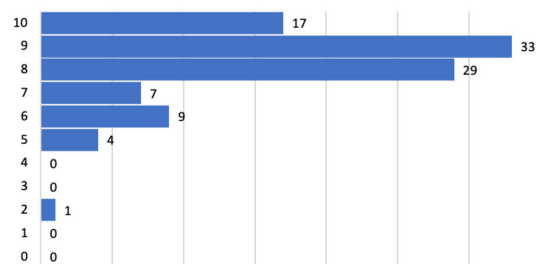

Figura 13. Resultados de las opiniones sobre la facilidad de uso del prototipo.

Respecto a la percepción asociada a la característica de claridad de configuración y ajustes del prototipo, la Figura 14 muestra los resultados de las opiniones las cuales indican que en la mayor concentración de ellas consideran estar totalmente de acuerdo (19), muy de acuerdo (30), de acuerdo (30), poco de acuerdo (19), muy poco de acuerdo (4) y opiniones neutras (2). La media en la escala de valores de las opiniones es de 9.22 con una desviación estándar de 1.177.

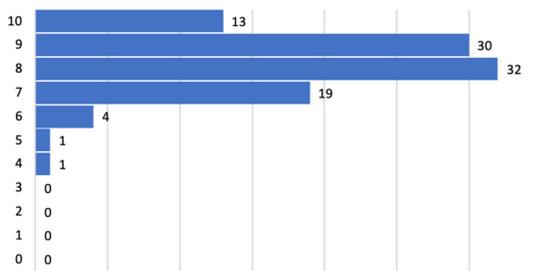

Figura 14. Resultados de las opiniones acerca de la claridad de la configuración y ajustes del prototipo.

El tercer aspecto evaluado muestra la percepción de los usuarios respecto a la facilidad de programación del prototipo. En este sentido, la Figura 15 muestra los resultados. Se observa la mayor cantidad de opiniones concentradas en las franjas de estar totalmente de acuerdo (16), muy de acuerdo (37), de acuerdo (34), seguido de poco de acuerdo (10) y un par de opiniones neutras (2) así como una opinión muy en desacuerdo (1). En este caso, la media para la escala de valores de las opiniones es de 9.44 y la desviación estándar es de 1.321.

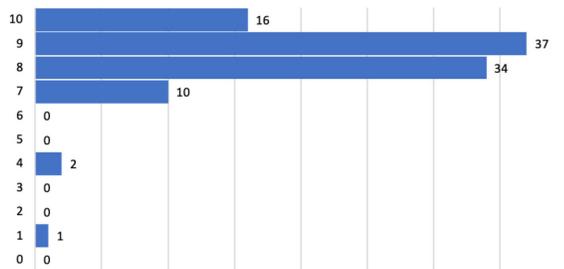

Figura 15. Resultados de las opiniones asociadas a la característica denominada facilidad de programación.

El último aspecto evaluado relacionado con la posibilidad de implementar aplicaciones futuras para el prototipo se muestra en la Figura 16. Se observa la mayor cantidad de opiniones concentradas en las franjas de estar totalmente de acuerdo (29), muy de acuerdo (38), de acuerdo (25), seguido de poco de acuerdo (3) y de cuatro opiniones neutras (2 y 2) seguidas de cero opiniones en contra. En este caso, la media para la escala de valores de las opiniones es de 9.84 y la desviación estándar es de 1.07

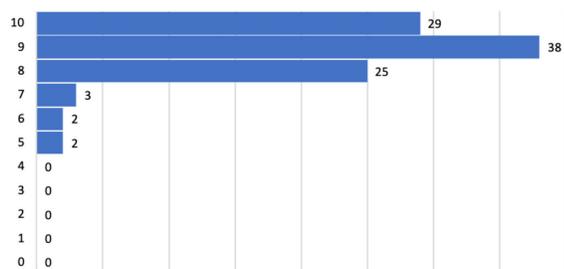

Figura 16. Resultados de las percepciones sobre la posibilidad de desarrollar aplicaciones futuras para el prototipo.

En resumen, las opiniones muestran que con una valoración promedio de 9.84, los encuestados consideran que el prototipo se puede adaptar fácilmente para su uso en distintos casos de estudio que se presenten en la industria. Por otra parte, con un valor promedio de 9.44, los usuarios consideraron que la interfaz de programación del prototipo es fácil e intuitiva, sin embargo, algunas opiniones consideran que se puede mejorar. Respecto a la claridad del procedimiento para configurar y ajustar el prototipo, los usuarios consideraron una nota promedio de valoración de 9.22 y finalmente, en cuanto al nivel de satisfacción asociado a la facilidad de uso del mecanismo se percibió una valoración promedio de 9.24.

## IV. CONCLUSIONES

En este trabajo se han descrito las características de construcción y funcionalidad de una herramienta para incentivar el desarrollo de competencias en el campo de los sistemas embebidos, de la visión artificial, así como de los aspectos que se deben de considerar para la solución de problemas reales. En los resultados que aquí se presentan se han utilizado tres casos de estudio que son muy comunes en la industria; identificación de códigos de barra, de elementos en PCBs y de tornillos. La evaluación del prototipo por parte de diferentes grupos de usuarios muestra que hay un gran interés en su uso y porque se siga trabajando en estos temas.

Las prácticas con el prototipo aportan ventajas significativas a los usuarios en aspectos de aprendizaje, capacitación y entrenamiento en los temas relacionados. Además, la facilidad de ajuste de los parámetros de operación del prototipo permite evaluar el desempeño del sistema en diferentes condiciones y determinar factores de influencia sobre la calidad de los resultados, dando lugar así a una mejor comprensión en el uso de estas tecnologías y en su uso para la solución apropiada de problemas reales. Esto representa entonces una herramienta que incentiva a los estudiantes en el uso de técnicas de sistemas embebidos, procesamiento de imágenes e inteligencia artificial para solucionar problemas industriales, académicos, sociales, ambientales, etc., así como en el uso de otros campos del conocimiento entre los que tenemos el control automático y la automatización.

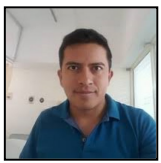
**Efrén Hernandez-Molina** nació en Tlaxcala, Tlaxcala, México en 1986. Obtuvo el grado de Ingeniero en Mecatrónica en la Universidad Politécnica de Tlaxcala. Actualmente es estudiante de la Maestría en Proyectos Tecnológicos de la Universidad de Guadalajara. Cuenta, además, con 10 años de experiencia laboral en la industria, especialmente en el área de ingeniería de proyectos. Sus temas de interés son: procesamiento de imágenes digitales, visión artificial y reconocimiento de patrones.

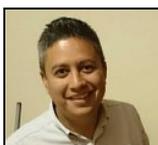
**Benjamín Ojeda-Magaña** nació en Guadalajara, Jalisco, México en 1976. En 2010 obtuvo el grado de Doctor en Tecnologías y Sistemas de Comunicaciones por la Universidad Politécnica de Madrid con reconocimiento "Cum Laude". En 2013 realizó un Posdoctorado en el Doctorado en Tecnologías de Información de la Universidad de Guadalajara. Ha publicado más de 30 artículos internacionales, tanto en congresos como en revistas especializadas, y ha participado como revisor en congresos y revistas científicas indizadas. Es miembro del Sistema Nacional de Investigadores (SNI) desde el 2014, actualmente es SNI I. En la actualidad es Profesor-Investigador en el Centro Universitario de Ciencias Exactas e Ingenierías (CUCEI) de la Universidad de Guadalajara. Sus áreas de interés son: procesamiento de imágenes digitales, reconocimiento de patrones, minería de datos, lógica difusa e inteligencia artificial.

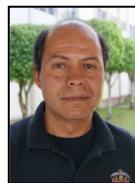
**José Guadalupe Robledo-Hernández** nació en Lagos de Moreno, Jalisco, en 1966. Obtuvo el grado de Ingeniero en Comunicaciones y Electrónica y el grado de Doctor en Tecnologías de Información por la Universidad de Guadalajara. Ha participado en proyectos de investigación con la Universidad de Cardiff, Gales, en el Reino Unido relacionados con prototipos rápidos. También ha colaborado en propulsar el Centro de Exploración de Soluciones para Ciudades Inteligentes IBM-UDG. Además, colaboró en la Dirección de Ciencia y Tecnología del Proyecto de Guadalajara Ciudad Creativa Digital en relación al desarrollo de la infraestructura tecnológica de la ciudad inteligente. Sus áreas de interés son: modelado y simulación de sistemas basados en multiagentes, uso de técnicas de analítica de datos bajo el enfoque de Big-data.

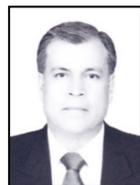
**Rubén Ruelas** obtuvo el grado de Ingeniero en Comunicaciones y Electrónica en 1988 por la Universidad de Guadalajara, un DEA en Control Automático y Procesamiento Digital de Señales en 1993 y el doctorado en Ingeniería Eléctrica en 1997, ambos grados por la Universidad Henri Poincaré-Nancy I, Francia, así como la especialidad en Tecnologías MEMS por la UNAM en el 2003. Actualmente es profesor en el Doctorado en Tecnologías de Información y en el posgrado en Ciencia de Materiales, todos ellos dentro del PNPC del CONACYT, es miembro del SNI y además lidera el grupo de Ingeniería de Manufactura dedicado a la investigación y desarrollo tecnológico en nanotecnología, celdas solares, fuentes de energías renovables, así como sistemas inteligentes e inteligencia artificial.